\documentclass[10pt,twocolumn,letterpaper]{article}

\usepackage[final]{cvpr}      
\usepackage{comment}
\usepackage{siunitx}
\usepackage{multirow}

\newcommand{\ad}[1]{\textcolor{red}{#1}}
\newcommand{\rh}[1]{\textcolor{blue}{#1}}
\definecolor{cvprblue}{rgb}{0.21,0.49,0.74}
\usepackage[pagebackref,breaklinks,colorlinks,allcolors=cvprblue]{hyperref}

\def\paperID{34467} 
\def\confName{CVPR}
\def\confYear{2026}

\title{Skullptor: High Fidelity 3D Head Reconstruction in Seconds \\
with Multi-View Normal Prediction}

\author{
    Noé Artru$^{1,2,3}$ \quad
    Rukhshanda Hussain$^{1,3}$ \quad
    Emeline Got$^{1}$ \quad
    Alexandre Messier$^{1}$ \quad \\
    David B.\ Lindell$^{2,4}$ \qquad
    Abdallah Dib$^{1}$\\ [7pt]
    $^{1}$Ubisoft La Forge \quad $^{2}$University of Toronto \quad $^{3}$ÉTS Montréal \quad $^{4}$Vector Institute
    \\ [7pt]
    {\href{https://ubisoft-laforge.github.io/character/skullptor/}{\color{cvprblue}\texttt{skullptor.github.io}}}
}

\begin{document}

\twocolumn[{
    \maketitle
    \begin{center}
    \vspace{-9pt}
    \includegraphics[width=\textwidth]{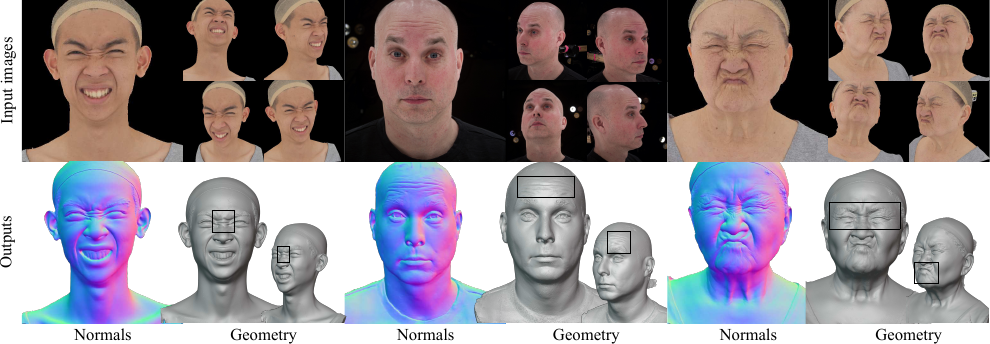}
    \end{center}
    \vspace{-12pt}
    \captionof{figure}{Given sparse, multi-view images of a subject, our method predicts geometrically consistent surface normals and leverages them to recover complete and detailed 3D head geometry, recovering fine surface features such as wrinkles and skin folds in a matter of seconds.}
    \label{fig:teaser}
    \vspace{7pt}
}]

\begin{abstract}
Reconstructing high-fidelity 3D head geometry from images is critical for a wide range of applications, yet existing methods face fundamental limitations. Traditional photogrammetry achieves exceptional detail but requires extensive camera arrays (25--200+ views), substantial computation, and manual cleanup in challenging areas like facial hair. Recent alternatives present a fundamental trade-off: foundation models enable efficient single-image reconstruction but lack fine geometric detail, while optimization-based methods achieve higher fidelity but require dense views and expensive computation. We bridge this gap with a hybrid approach that combines the strengths of both paradigms. Our method introduces a multi-view surface normal prediction model that extends monocular foundation models with cross-view attention to produce geometrically consistent normals in a feed-forward pass. We then leverage these predictions as strong geometric priors within an inverse rendering optimization framework to recover high-frequency surface details. Our approach outperforms state-of-the-art single-image and multi-view methods, achieving high-fidelity reconstruction on par with dense-view photogrammetry while reducing camera requirements and computational cost. Results available on the \href{https://ubisoft-laforge.github.io/character/skullptor/}{project page}.
%

\end{abstract}    
\section{Introduction}
Obtaining high-fidelity 3D facial geometry from images is a fundamental challenge with critical importance across numerous applications, from visual effects and gaming to virtual communication. Geometric accuracy in fine details such as skin texture and subtle surface variations directly impacts user experience and is key to bridging the uncanny valley.

Photogrammetry \cite{hartley2003multiple, szeliski2022computer, furukawa2015multi} has long served as the gold standard for acquiring detailed 360-degree head geometry in production pipelines, making it the preferred choice for high-end VFX and gaming applications. However, it requires dense camera coverage (typically 50--200+ synchronized views), extensive processing time, and substantial computational resources. For 4D sequences, the storage requirements become prohibitive with multi-view capture, generating terabytes of data. Moreover, photogrammetry reconstruction often struggles with the presence of view-dependent appearance (e.g., specular reflections) and fine geometry such as facial hair, leading to artifacts and incomplete geometry that require manual correction.

Recently, two technical approaches have emerged to reduce the processing time and the number of viewpoints required for high-fidelity 3D reconstruction; however, both navigate a fundamental trade-off between speed, ease of data capture, and geometric accuracy.
The first leverages data-driven foundation models for human vision~\cite{khirodkar2024sapiens, saleh2025david, oquab2023dinov2, wei2024meshlrm}, which recover 3D geometry from a single image in an efficient feed-forward pass, dramatically simplifying data capture compared to photogrammetry.
Still, even state-of-the-art foundation models~\cite{khirodkar2024sapiens} fall short of photogrammetry in reconstructing fine geometric detail, likely because they rely on learned, and often ambiguous, 3D shape priors rather than precise multi-view geometric constraints.

A second approach reconstructs 3D geometry through direct, iterative optimization of implicit or explicit 3D representations~\cite{guedon2024sugar, Huang2DGS2024, rosu2023permutosdf}.
These representations are easier to optimize than traditional mesh-based formulations and achieve higher geometric fidelity than foundation models by explicitly enforcing multi-view consistency, yet they still struggle to capture high-frequency geometric details such as wrinkles, skin folds, and person-specific surface variations.
Moreover, because they lack strong data-driven priors, they typically require extensive camera arrays to provide dense view coverage, as well as computationally expensive optimization procedures.
Consequently, this class of methods lags behind foundation models in terms of computational efficiency and ease of capture.
Overall, neither photogrammetry nor these emerging methods achieve all three desiderata simultaneously: high geometric accuracy, a sparse view capture setup, and computationally efficient reconstruction.

Here, we address this gap by combining the complementary strengths of foundation models and direct, iterative optimization techniques for production-quality facial capture. 
Specifically, we propose a novel approach that achieves dense-view, photogrammetry-level fidelity from sparse camera views (fewer than ten) through two key innovations.
First, we introduce a multi-view surface normal prediction model that produces geometrically consistent normals from sparse viewpoints in an efficient, feed-forward fashion.
Our normal estimator builds on a monocular foundation model \cite{saleh2025david} trained on synthetic facial images and extends it to enforce multi-view consistency by incorporating multi-head attention across the input viewpoints. 
Second, we present a direct optimization framework based on inverse rendering \cite{palfinger2022continuous} that leverages the predicted normals as strong geometric priors to refine the 3D reconstruction and recover high-frequency surface details.
Together, these innovations substantially reduce the required camera coverage and computational cost relative to traditional photogrammetry and pure optimization-based methods. Specifically, our approach reconstructs a highly detailed, complete 3D head from 10 cameras in 30 seconds,
achieving photogrammetry-level quality (\autoref{fig:teaser} and \autoref{fig:geom_comp}). Moreover, our method scales gracefully to even sparser setups, enabling high-fidelity frontal facial reconstruction with as few as three cameras.

In summary, we make the following contributions.
\begin{itemize}
\item We develop a multi-view normal prediction model that efficiently adapts a monocular facial foundation model (DAViD~\cite{saleh2025david}) using a lightweight view-aware cross-attention mechanism to enforce geometric consistency, generating significantly more accurate surface normals from sparse ($<$10) input viewpoints.
%
%
\item We leverage our data-driven, feed-forward prediction of surface normals within an inverse rendering optimization framework to recover high-frequency surface details.
\item We demonstrate that the proposed approach outperforms state-of-the-art dense multi-view and single-image approaches while achieving dense-view photogrammetry fidelity with reduced camera numbers and processing time.
\end{itemize}
We release our code and model to facilitate future research.
\section{Related Work}
Multi-view stereo (photogrammetry) \cite{hartley2003multiple, szeliski2022computer, furukawa2015multi}, implemented in open frameworks like COLMAP \cite{schoenberger2016sfm, schoenberger2016mvs} and Meshroom \cite{Meshroom}, as well as commercial solutions\footnote{%
\begin{minipage}[t]{\linewidth}
RealityCapture: \url{https://www.capturingreality.com}\\
Agisoft: \url{https://www.agisoft.com}\end{minipage}}, remains the gold standard for 3D facial capture in production. These methods achieve exceptional accuracy by triangulating 3D structure from dense correspondences across multiple views. However, photogrammetry requires dense camera coverage (25--250+ views), substantial computational resources, and struggles with non-Lambertian surfaces and fine structures like facial hair. Recent approaches aim to simplify multi-view reconstruction through learned priors or flexible 3D representations, yet navigate a fundamental trade-off between geometric accuracy, computational efficiency, and capture requirements.
\\
\textbf{Data-driven methods.} Data-driven approaches prioritize computational efficiency by leveraging learned priors from large-scale datasets, typically trained on monocular 2D images. Although fast, these methods typically sacrifice geometric accuracy due to the inherent ambiguity in mapping 2D observations to 3D structure without explicit multi-view constraints.

\emph{Parametric face models}, particularly 3D Morphable Models (3DMM) \cite{blanz1999morphable, li2017learning, Egger20Years}, model faces as linear combinations of basis shapes learned from scan databases. Methods \cite{feng2021learning, danvevcek2022emoca, dib2021towards, chai2023hiface, lei2023hierarchical, dib2022s2f2, dib2021practical, wang20243d} fit these models to images through optimization or regression, with recent works using graph neural networks \cite{lin2020towards, gao2020semi, lee2020uncertainty, Dib_2024_CVPR} for refinement. While computationally efficient, parametric approaches cannot represent details beyond their training data, are typically restricted to frontal regions, and struggle with person-specific features outside the learned distribution.

\emph{Monocular normal and depth estimation models} offer greater flexibility than parametric models by predicting per-pixel geometric information without topological constraints. Recent estimators achieve strong performance through pre-training on domain-specific synthetic data \cite{saleh2025david, khirodkar2024sapiens}, leveraging large-scale diverse datasets \cite{depth_anything_v1, depth_anything_v2}, or by fine-tuning general-purpose models \cite{ye2024stablenormal} like Marigold \cite{ke2025marigold} (adapted from Stable Diffusion) and Pixel3DMM \cite{giebenhain2025pixel3dmm} (built on DINOv2 \cite{oquab2023dinov2}). DAViD \cite{saleh2025david} uses Dense Prediction Transformers \cite{ranftl2021visiontransformersdenseprediction}, combining vision transformers \cite{dosovitskiy2021imageworth16x16words} with convolutional blocks for high-resolution feature extraction. However, monocular estimators suffer from inherent single-image ambiguities and process each view independently, so geometric consistency across viewpoints is not guaranteed, leading to contradictory predictions that cannot be directly integrated into a coherent 3D mesh reconstruction.
%

\emph{Mesh generation models} represent the most flexible class of data-driven approaches. Multi-view diffusion models like Zero-1-to-3 and Zero123++ \cite{liu2023zero1to3, shi2023zero123plus} synthesize novel views for reconstruction, while feed-forward models such as Unique3D \cite{wu2024unique3dhighqualityefficient3d}, PSHuman \cite{li2025pshumanphotorealisticsingleimage3d}, CraftsMan \cite{li2024craftsman}, CAP4D \cite{taubner2024cap4dcreatinganimatable4d}, and MeshLRM \cite{wei2024meshlrm} directly produce 3D meshes or implicit representations from single or sparse views. While fast and capable of producing plausible detail, these methods rely on learned priors to hallucinate geometry rather than enforcing explicit multi-view constraints, fundamentally limiting their ability to capture person-specific fine details and achieve the geometric accuracy required for high-fidelity facial reconstruction.
%
\\
\textbf{Optimization-based methods.} Direct optimization methods reconstruct 3D geometry through iterative refinement of explicit or implicit representations. Early approaches use voxel grids \cite{jackson2017large} or meshes \cite{cheng2020faster, kanazawa2018learning, wang2018pixel2mesh}, but are limited by fixed topologies and memory constraints. Recent methods leverage flexible implicit representations such as signed distance functions \cite{rosu2023permutosdf, ren2023facial, guo2023rafare} or combine reflectance and normal maps within volume rendering frameworks \cite{Bruneau25, Brument24}. Others explore explicit point-based representations, particularly 3D Gaussian Splatting \cite{kerbl3Dgaussians}, with extensions like SuGaR \cite{guedon2024sugar} and 2DGS \cite{Huang2DGS2024} that extract surface meshes. Ramon et al.~\cite{ramon2021h3d} use a prior head model from thousands of scans to overfit a neural network via differentiable rendering \cite{niemeyer2020differentiable}. These methods achieve higher geometric fidelity than data-driven approaches by explicitly enforcing multi-view photometric consistency. However, without strong learned priors, they typically require dense view coverage and computationally expensive per-scene optimization, making them impractical for sparse-view scenarios.
%


\noindent\textbf{Inverse rendering.} Differentiable rendering libraries such as \cite{ravi2020accelerating3ddeeplearning, Laine2020diffrast} enable inverse rendering by optimizing mesh geometry and appearance to match observed images. This approach has a long history, with foundational work establishing energy functions that balance data fidelity with geometric regularization \cite{10.1145/166117.166119, 10.1145/1057432.1057457, :10.2312/SGP/SGP03/020-030, 10.1145/2816795.2818078}. However, directly applying gradient updates with standard optimizers like ADAM \cite{kingma2017adammethodstochasticoptimization} often leads to mesh degradation, including self-intersections and flipped triangles. While periodic remeshing \cite{botsch2010polygon, alliez2008recent} can mitigate these issues, it disrupts the optimization process. Advanced techniques such as Continuous Remeshing \cite{palfinger2022continuous} integrate adaptive topology management directly into each optimization iteration, preventing mesh defects and enabling stable refinement of high-frequency geometric details.
\vspace{-10px}
\paragraph{Our approach.} Data-driven methods achieve computational efficiency through learned priors but sacrifice geometric accuracy due to the absence of explicit multi-view constraints. Conversely, optimization-based methods achieve higher geometric fidelity by enforcing multi-view photometric consistency, but without strong learned priors, they require dense view coverage and computationally expensive per-scene optimization. We bridge this gap by combining data-driven multi-view normal prediction with inverse rendering optimization.
\section{Reconstructing 3D Head Meshes in Seconds}
\label{sec:method}

\begin{figure*}[ht]
    \centering
    \includegraphics[width=1\linewidth]{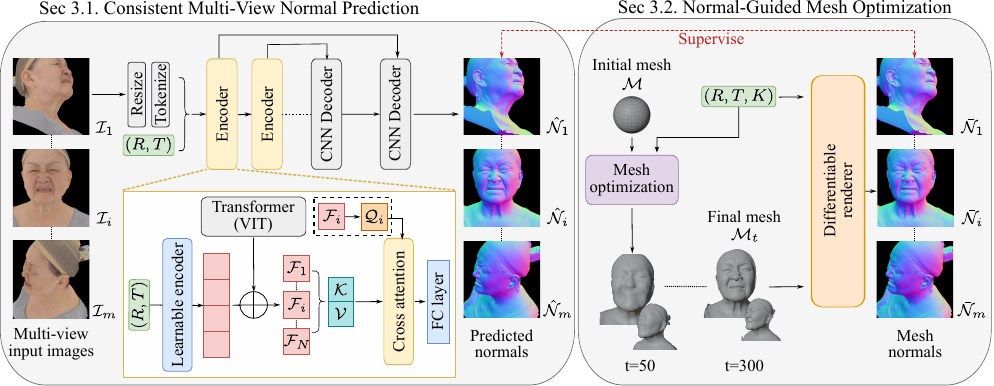}
    \caption{Skullptor reconstructs 3D meshes in two stages. Multi-view normal prediction (Sec.\ 3.1) produces geometrically consistent surface normals from sparse input images by leveraging cross-view attention across all viewpoints. Mesh optimization (Sec.\ 3.2) then refines the 3D geometry using the predicted normals as geometric priors within an inverse rendering framework.}
    \label{fig:diag_method}
    \vspace{-10pt}
\end{figure*}
Our framework decomposes 3D facial mesh reconstruction into two stages.
In the first stage, it estimates normal maps from sparse multi-view images, using a multi-view transformer architecture to fuse information across viewpoints. 
In the second stage, the predicted normal maps from the first stage are used to reconstruct the 3D head through an inverse rendering optimization. 
This formulation enables efficient and high-fidelity geometry reconstruction while avoiding the computational overhead of traditional photogrammetry pipelines.
An overview of the pipeline is in \autoref{fig:diag_method}.
%

\subsection{Consistent Multi-View Normal Prediction}
\label{method_normal}
Given a set of $m$ multi-view color images $\mathcal{I} = \{\mathcal{I}_1, \mathcal{I}_2, \dots, \mathcal{I}_m\}$  and their corresponding camera parameters $\{(R_i, T_i, K_i)\}_{i=1}^m$, where $R_i \in \mathbb{R}^{3\times3}$,  $T_i \in \mathbb{R}^3$, and $K_i \in \mathbb{R}^{3\times3}$  denote the rotation, translation, and intrinsic matrices respectively, our goal is to predict a set of normal maps  $\mathcal{N} = \{\mathcal{N}_1, \mathcal{N}_2, \dots, \mathcal{N}_m\}$  consistent across all views. 
Formally, we seek a model $\mathbf{\Psi}$ that performs multi-view per-pixel normal estimation:
\begin{equation}
\mathcal{N} = \mathbf{\Psi}(\mathcal{I}, \{(R_i, T_i)\}_{i=1}^m).
\end{equation}

We base our multi-view normal predictor on the DAViD foundation model~\cite{saleh2025david}, originally designed for dense single-view normal estimation and trained on a large corpus of synthetic data. DAViD expands upon the Dense Prediction Transformer architecture \cite{ranftl2021visiontransformersdenseprediction}, making it particularly well-suited for per-pixel prediction tasks by combining the global context modeling of vision transformers with the high-resolution processing capabilities of CNNs.
%

We modify the DAViD architecture to enforce multi-view consistency and propagate high-frequency geometric cues across views. For this, we introduce view-aware cross-attention layers within each transformer encoder block of the original DAViD architecture. In our context, cross-attention allows each image to incorporate information from all other viewpoints when predicting the corresponding normal map. 

Concretely, let the encoded feature sequence for the $i$-th image be $\mathcal{F}_i \in \mathbb{R}^{L \times D}$, where $L = 577$ is the token length and $D = 1024$ is the feature dimension. $\mathcal{F}_i$ are obtained by splitting input images into patches and passing each patch through a patch embedding layer. We insert a multi-view cross-attention layer between each transformer block and fully connected layer. For the target view $i$, its query $\mathcal{Q}_i$ attends to the keys and values constructed from all views:
\begin{align}
\mathcal{Q}_i &= \mathcal{F}_i \mathcal{W}_Q, \\
\mathcal{K}   &= [\mathcal{F}_1 \mathcal{W}_K; \mathcal{F}_2 \mathcal{W}_K; \dots; \mathcal{F}_m \mathcal{W}_K], \\
\mathcal{V}   &= [\mathcal{F}_1 \mathcal{W}_V; \mathcal{F}_2 \mathcal{W}_V; \dots; \mathcal{F}_m \mathcal{W}_V],
\end{align}

\noindent
where $\mathcal{W}_Q, \mathcal{W}_K, \mathcal{W}_V \in \mathbb{R}^{D \times D}$ are learnable projection matrices, and $[\,;\,]$ denotes concatenation along the sequence dimension. The attention output is computed as:
\[
\text{Attention}(\mathcal{Q}_i, \mathcal{K}, \mathcal{V}) = \text{softmax}\left(\frac{\mathcal{Q}_i \mathcal{K}^\top}{\sqrt{D}}\right)\mathcal{V}.
\]

Additionally, we encode each camera transformation as a positional embedding derived from its extrinsic parameters $(R_i, T_i)$.  
We first convert the rotation matrix $R_i$ into a unit quaternion  
$\mathbf{q}_i = (q_w, q_x, q_y, q_z)^\top$,  
forming a 7-dimensional camera pose representation  
$\mathbf{p}_i = [\mathbf{q}_i; \mathbf{T}_i] \in \mathbb{R}^7$.  
This pose vector is projected to the feature dimension and added to each token in  $\mathcal{F}_i$ before computing cross-attention,  
enabling the model to distinguish tokens originating from different viewpoints more effectively.
\\
\textbf{Training.}
The model takes $m$ input images $\mathcal{I}$ and predicts their corresponding normal maps $\hat{\mathcal{N}}$ at the same spatial resolution. We use cosine similarity loss as the objective function:
\begin{equation}
\mathcal{L}_{\text{cos}} = 1 - \frac{1}{m} \sum_{i=1}^{m} \mathcal{\hat{N}}_i \cdot \mathcal{N}_i,
\end{equation}
where $\hat{\mathcal{N}}_{i}$ is the predicted normal for the $i^{\text{th}}$ view  and $\mathcal{N}_{i}$ is the corresponding ground truth normal.

\begin{table*}[h!]
\centering
\caption{Normal estimation comparison on Multiface and NPHM datasets.}
\label{tab:normal_comp_merged}
\footnotesize
\begin{tabular}{@{}llccccccc@{}}
\toprule
Dataset & Method & Avg. Angular Error $\downarrow$  & Avg. Normal Grad. Err. $\downarrow$ & $<\ang{10}$ (\%) $\uparrow$ & $<\ang{20}$ (\%) $\uparrow$ & $<\ang{30}$ (\%) $\uparrow$ & Time (s) $\downarrow$ \\
\midrule
\multirow{4}{*}{\rotatebox[origin=c]{90}{Multiface}} 
& Sapiens 0.3B & 10.3 & 0.265 & 61.9 & 85.3 & \underline{94.6} & 7.2 \\
& Sapiens 2B & 9.23 & 0.257 & \underline{69.3} & \underline{86.3} & 94.2 & 41.3\\
& DAViD & \underline{9.16} & \underline{0.250} & \textbf{69.7} & \textbf{86.7} & \textbf{94.7} & \textbf{1.1} \\
\cmidrule{2-8}
& Skullptor (Ours)  & \textbf{9.13} & \textbf{0.234} & 69.2 & \textbf{86.7} & \underline{94.6} & \underline{1.5} \\
\midrule\midrule
\multirow{4}{*}{\rotatebox[origin=c]{90}{NPHM}} 
& Sapiens 0.3B & 7.33 & 0.194 & \underline{78.4} & 93.4 & 97.8 & 8.2 \\
& Sapiens 2B & \textbf{6.86} & \underline{0.185} & \textbf{80.0} & \textbf{94.4} & \textbf{98.3} & 43.7 \\
& DAViD & 7.86 & 0.190 & 73.8 & 92.5 & \underline{98.1} & \textbf{1.1} \\
\cmidrule{2-8}
& Skullptor (Ours) & \underline{7.29} & \textbf{0.166} & 76.8 & \underline{94.1} & \textbf{98.3} & \underline{1.5} \\
\bottomrule
\end{tabular}%
\vspace{-10pt}
\end{table*}

For training, we initialize the layers corresponding to DAViD with pre-trained DAViD weights to leverage the knowledge from its original training dataset, and then fine-tune the entire model on a dataset comprised of high quality 3D head scans from Triplegangers \cite{triplegangers}. We selected a diverse sample of 50 subjects, each with 20 static expressions captured from 55 lightstage cameras. The Triplegangers scans are obtained using photogrammetric reconstruction with subsequent manual cleanup to correct artifacts and ensure high geometric fidelity. 5 subjects were left out of the training dataset for validation and visualization purposes.

Rather than training on images from fixed lightstage camera positions, we render the ground truth geometry (with its corresponding texture) from synthesized viewpoints with randomly sampled virtual cameras.  For each training sample, we render both the textured image $\mathcal{I}_i$  and its corresponding ground truth normal map $\mathcal{N}_i$ (in camera space) from the head geometry. This rendering-based data augmentation provides two key advantages: (1) it exposes the model to diverse camera configurations beyond the lightstage setup, improving generalization to arbitrary multi-view capture systems, and (2) it increases training data diversity without requiring additional physical captures. As we demonstrate in Sec.\ \ref{sec:evaluation}, our model is able to generalize effectively to different capture setups, including NPHM and Multiface datasets with different camera configurations. Technical details are provided in the supplementary material.

\subsection{Normal-Guided Mesh Optimization}

\label{ssec:mesh_optim}

Given the predicted multi-view surface normals $\hat{\mathcal{N}}$ (Sec.\ \ref{method_normal}), we recover detailed 3D facial geometry through optimization-based mesh refinement. Our approach iteratively adjusts mesh vertex positions such that the rendered surface normals match the predicted normals under differentiable rendering. A key challenge is establishing geometric correspondence between the predicted normals, expressed in local camera coordinates, and a consistent global coordinate space.
\\
\textbf{Coordinate system calibration.}
To enable consistent optimization across multi-view captures, we align the predicted normals and camera parameters to a canonical coordinate system defined by a template mesh 
with fixed topology normalized to unit radius. We first detect 2D facial landmarks~\cite{zheng2022farl} in each input image $\hat{\mathcal{I}}_i$ and triangulate them into 3D by finding points that minimize re-projection error across all views via least-squares ray intersection. Given the resulting landmark set $\mathbf{X}$ and corresponding template landmarks $\mathbf{Y}$, we compute the optimal similarity transformation $\mathcal{G} = \{R_g, T_g, S_g\}$ that aligns $\mathbf{X}$ to $\mathbf{Y}$ using Procrustes analysis \cite{arun1987least} with singular value decomposition.  

Next, we apply $\mathcal{G}^{-1}$ to the camera extrinsics, yielding calibrated transformation: $(R_i^*, T_i^*)  = G \circ (R_i, T_i)$. We then transform the predicted normals from camera space to the canonical space via rotation: $  \hat{\mathcal{N}}^* = R^{-1}_{g} \cdot \hat{\mathcal{N}}$. We additionally mask the normal maps using DAViD's foreground estimation model~\cite{saleh2025david} to exclude background regions.
\\
\textbf{Mesh optimization.}
We initialize the mesh $\mathcal{M}$, that we want to optimize, to a unit sphere, and optimize it to minimize the distance between the predicted normals  $\hat{\mathcal{N}}^*$ (from previous stage) and the normals rendered $\bar{\mathcal{N}}$ from the $\mathcal{M}$. At each step of the optimization, we render the mesh under the transformed camera parameters using a differentiable rendering function $\mathcal{F}$ defined as follows: 
\begin{align}
\bar{\mathcal{N}}_{i} = \mathcal{F}(\mathcal{M}, R_i^{*}, T_i^{*}, K_i), \quad i = 1, \ldots, m.
\end{align}
We optimize the mesh to minimize the objective function
\begin{equation}
\mathcal{L} = \mathcal{L}_{\mathrm{normal}}(\bar{\mathcal{N}}, \mathcal{N}^*) + \lambda_{\mathrm{lap}} \mathcal{L}_{\mathrm{lap}}.
\end{equation}
The normal loss measures cosine similarity between rendered and predicted normals across all views:
\begin{equation}
\mathcal{L}_{\mathrm{normal}} = 1 - \frac{1}{m} \sum_{i=1}^{m} \hat{\mathcal{W}}_i  \cdot ( \mathcal{\hat{N}^*}_i \cdot \bar{\mathcal{N}}_i),
\end{equation}
where $\hat{\mathcal{W}}_{i}$ is a per-pixel weight matrix for the view $i$. 
This weight matrix prioritizes regions facing the camera where predictions are more reliable, defined at a given pixel position $(u, v)$ by 
\begin{equation}
\hat{\mathcal{W}}_{i}(u,v) = \frac{ \exp\!\left[\, \alpha \big( \hat{\mathcal{N}}_{i}^{*}(u,v) \cdot \boldsymbol{d_i} \big) \right] - 1 }
     { \exp(\alpha) - 1 }.
\end{equation}
%
%
%
Here, $\boldsymbol{d_i}$ is the camera viewing direction and $\alpha$ controls how strongly frontal-facing regions are prioritized over grazing angles. 
Finally, the Laplacian regularization term $\mathcal{L}_{\mathrm{lap}}$ encourages local smoothness~\cite{nealen2006laplacian} and $\lambda_{\mathrm{lap}}$ its weight.

At the end of each optimization step, we apply adaptive remeshing operations from ~\cite{palfinger2022continuous}. These operations consist of edge splits, collapses, and flips that dynamically adjust mesh resolution to local geometric complexity while preventing degeneracies such as collapsed faces and self-intersections. The final output is a normalized mesh aligned to the canonical coordinate system. While we initialize $\mathcal{M}$ with a unit sphere for simplicity, using an aligned template head mesh as initialization did not show meaningful differences in terms of reconstruction time or quality. More details are available in the supplemental material.

\section{Evaluation}
\label{sec:evaluation}

The evaluation of our proposed pipeline aims to validate three key claims: 
(1) our extension of monocular normal estimation models to multi-view processing improves normal map estimation, especially for capturing high-frequency details; 
(2) our optimization-based remeshing effectively leverages these multi-view normal estimates to produce a detailed mesh; and
(3) the complete pipeline outperforms standard photogrammetry in quality when both methods operate on sparse camera arrays ($< 10$), while also being more computationally efficient.

\subsection{Experimental Setup}
\label{subsec:setup}

\begin{figure}[t]
    \centering
    \includegraphics[width=1\linewidth]{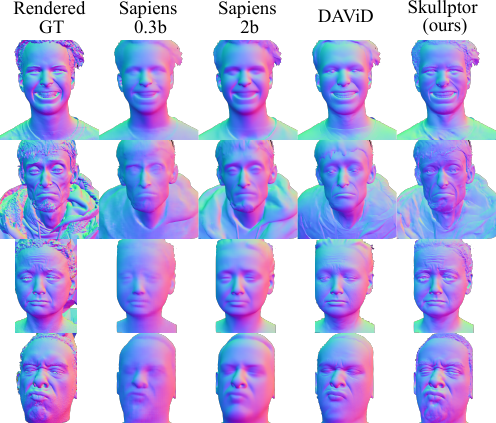}
    \caption{Qualitative comparison of normal predictions on the NPHM dataset (rows 1--2) and Multiface dataset (rows 3--4).}
    \label{fig:normal_comp}
  \vspace{-15pt}
\end{figure}

    \paragraph{Datasets.} We evaluate our approach on two publicly available datasets: NPHM \cite{giebenhain2023nphm} and Multiface \cite{wuu2023multifacedatasetneuralface}.
    \\
    \textit{NPHM.} This dataset contains highly detailed 3D head scans of 124 subjects, each performing 20 distinct facial expressions, captured with a custom structured-light scanning setup using Artec Eva scanners~\cite{EvaScanner}. This capture setup provides highly accurate geometry estimations without using a photogrammetry pipeline, making it ideal for evaluations involving photogrammetry. The dataset does not contain captured images, so we generate a synthetic dataset by rendering textured NPHM scans from 23 virtual camera viewpoints. We render normal maps and textured images to serve as ground truth and input data, respectively. For a tractable evaluation, we use a subset consisting of the first 20 subjects, for a total of 400 expression scans.
    \\
    \textit{Multiface.} This dataset is composed of lightstage multi-view video recordings of 8 subjects from 40 different cameras performing a variety of facial expressions, resulting in thousands of dynamic frames. For this dataset, we directly use the captured images as input. Ground-truth geometry is obtained via a high-quality photogrammetry reconstruction pipeline, producing dense, high-fidelity meshes for each frame. To match the scale of our NPHM dataset, we subsample the sequences by selecting 1 out of every 30 frames, yielding a diverse evaluation set of 347 frames in total.
    
    For both datasets, all ground-truth meshes are aligned to a canonical coordinate system using the same technique described in Sec.~\ref{ssec:mesh_optim}. This ensures all reconstructions are in a canonical space, allowing for consistent results on metrics that depend on scale, such as depth error. Further details on how the evaluation dataset was created are available in the supplemental material.
    \\
    \textbf{Baselines.}
    We compare our method against several state-of-the-art baselines at two different stages:
    
    \begin{itemize}
        \item \textbf{Normal estimation.} We compare our multi-view normal estimator against leading monocular methods, specifically, the foundation model Sapiens \cite{khirodkar2024sapiens} using both its lightweight 0.3B architecture, as well as its full 2B parameter mode, and the original DAViD model \cite{saleh2025david}.
        
        \item \textbf{Full mesh reconstruction.} We compare our complete pipeline against established multi-view reconstruction methods: Meshroom \cite{Meshroom} (a robust open-source photogrammetry pipeline) and recent Gaussian splatting-based methods 2DGS \cite{Huang2DGS2024} and SuGaR \cite{guedon2024sugar}, which recover explicit geometry from Gaussian splats.
    \end{itemize}
    \textbf{Evaluation metrics.} We evaluate normal prediction quality using: (1) \emph{Mean Angular Difference} (in degrees), measuring the angular error between predicted and ground truth normals; (2) \emph{Percentage of angles below $\ang{10}$, $\ang{20}$, and $\ang{30}$}, measuring the fraction of predictions whose angular error falls within these thresholds; and (3) \emph{Normal Gradient Error}, a novel metric that captures preservation of high-frequency geometric details. This metric applies a Sobel filter to both predicted and ground truth normals and measures the L1 distance between filtered results. This metric directly measures the preservation of fine geometric details by capturing local surface curvature variations. Since normals are spatial derivatives of depth, their gradients emphasize high-frequency features such as wrinkles, folds and subtle surface variations that are critical for photorealistic reconstruction. Methods that produce overly smooth geometry may achieve low angular error while scoring poorly on normal gradient error, making this metric particularly relevant for evaluating our approach's ability to capture fine-scale detail.
    Additionally, for mesh evaluations, we include an L1 depth error to assess the global quality of the mesh reconstruction.

\begin{table*}[t]
\centering
\caption{Mesh reconstruction accuracy comparison on Multiface and NPHM datasets.}
\label{tab:geom_comp_merged}
\footnotesize
\begin{tabular*}{\textwidth}{@{\extracolsep{\fill}} llcccccccc @{}}\toprule
\multirow{2}{*}{Dataset} & \multirow{2}{*}{Method} & Depth Error & Avg. Angular & Avg. Normal & $<\ang{10}$ & $<\ang{20}$ & $<\ang{30}$ & Runtime & \# Views \\
& & (mm) $\downarrow$ & Error $\downarrow$ & Gradient Error $\downarrow$ & (\%) $\uparrow$ & (\%) $\uparrow$ & (\%) $\uparrow$ & (min) $\downarrow$ & \\
\midrule
\multirow{4}{*}{\rotatebox[origin=c]{90}{Multiface*}} 
& Meshroom (Photogrammetry) & \textbf{0.467} & \textbf{5.43} & \textbf{0.143} & 85.1 & \underline{95.8} & \textbf{99.0} & 7.8  & 26 \\
& 2D GS & 5.73 & 10.7 & 0.206 & 68.6 & 85.4 & 92.5 & 50  & 26 \\
& SuGaR & 5.54 & 11.5 & 0.324 & 66.1 & 83.8 & 91.1 & 42  & 26 \\
\cmidrule{2-10}
& Skullptor (Ours) & \underline{2.43} & \underline{6.17} & \underline{0.156} & \textbf{91.5} & \textbf{97.6} & \textbf{99.0} & \underline{0.67} & 26 \\
& Skullptor (Ours, 10 Views) & 2.99 & 6.46 & 0.157 & \underline{86.8} & \underline{95.8} & \underline{97.9} & \textbf{0.48}& 10 \\
\midrule\midrule
\multirow{5}{*}{\rotatebox[origin=c]{90}{NPHM*}} 
& Meshroom (Photogrammetry) & 2.54 & \textbf{5.25} & \textbf{0.114} & \textbf{88.3} & 96.0 & \underline{98.0} & 9.5 & 23 \\
& 2D GS & 6.37 & 6.64 & 0.134 & 84.3 & 93.2 & 95.8 & 47 & 23 \\
& SuGaR & 3.23 & 9.03 & 0.232 & 77.9 & 91.3 & 95.6 & 50 & 23 \\
\cmidrule{2-10}
& Skullptor (Ours) & \textbf{2.33} & \underline{6.10} & \underline{0.115} & \underline{87.3} & \textbf{96.3} & \textbf{98.1} & \underline{0.72} & 23 \\
& Skullptor (Ours, 10 Views) & \underline{2.36} & 6.12 & 0.117 & \underline{87.3} & \underline{96.2} & \textbf{98.1} & \textbf{0.50} & 10 \\
\bottomrule
\end{tabular*}

\captionsetup{justification=justified, singlelinecheck=false}
\caption*{\footnotesize\textbf{*} Metrics exclude 6 Multiface and 12 NPHM subjects from the initial 50 faces per dataset where 2DGS or SuGaR failed to produce a reconstruction.}
\vspace{-15pt}
\end{table*}

\subsection{Normal Evaluation}
\label{subsec:normal_evaluation}

In this section, we validate the effectiveness of our multi-view normal estimation network (Sec.~\ref{method_normal}). We compare it against the monocular baselines defined in Sec.~\ref{subsec:setup}. For ground truth, we rasterize normals from the ground truth meshes with camera views identical to the ones used for input images. Additionally, we use the Facer face segmentation~\cite{zheng2022farl} to filter out hair, cloth and neck of each subject, ensuring that normal prediction metrics are calculated only on the face. We also report the average inference time for predicting one batch of 10 images.

As shown in Table~\ref{tab:normal_comp_merged}, monocular methods (Sapiens, DAViD) achieve competitive angular errors, but these metrics do not heavily penalize errors on fine-scale geometric details, unlike the normal gradient error. On this metric, the monocular methods perform worse, reflecting their loss of person-specific visual details (see \autoref{fig:normal_comp}). Our model recovers these details while introducing minimal computational overhead ($1.5$ sec) as compared to DAViD, remaining an order of magnitude faster than Sapiens 2B.
\begin{figure}[b]
    \centering
    \vspace{-5pt}
    \includegraphics[width=1\linewidth]{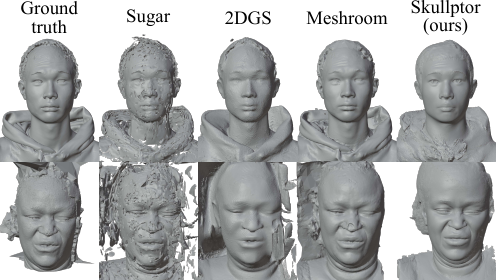}
    \caption{Qualitative comparison of mesh reconstruction methods on the NPHM (top) and Multiface (bottom) datasets.}
    \vspace{0pt}
    \label{fig:geom_comp}
\end{figure}
\subsection{Geometry Evaluation}
\label{subsec:pipeline_comparison}
We now compare our full pipeline, which produces meshes from sets of multi-view image, against other reconstruction baselines (Meshroom \cite{Meshroom}, 2DGS \cite{Huang2DGS2024}, SuGaR \cite{guedon2024sugar}). 
We first align each predicted mesh to the ground truth mesh using Procrustes analysis as described in Sec.~\ref{ssec:mesh_optim}.
For evaluation, we rasterize depth and normal maps from 12 novel camera views (unseen during reconstruction) at 512$\times$512 resolution for each predicted and ground truth mesh. We also render the textured ground truth mesh in order to estimate a facial mask using Facer~\cite{zheng2022farl}, to evaluate the geometry of the face region.

To ensure a representative yet computationally feasible evaluation, we limit this comparison to a randomly chosen subset of 50 expressions/frames from each dataset (NPHM and Multiface), for a total of 100 facial reconstructions. Quantitative results are summarized in Table~\ref{tab:geom_comp_merged}.  

Our approach substantially outperforms 2DGS and SuGaR across all metrics on both datasets while being an order of magnitude faster and using significantly fewer camera views. Compared to  Meshroom, our method achieves on-par geometric quality on NPHM while using less than half the input views (10 vs.\ 23) and being an order of magnitude faster.  

Note that the Multiface ground truth (40 views) and Meshroom baseline (26 views) are both photogrammetry-based, which may lead to correlated biases for Meshroom on Multiface. For instance, the ‘broken nose’ artifact in the Multiface photogrammetry (GT+Meshroom, \autoref{fig:geom_comp}), seems to penalize our correct reconstruction. In this specific case, our method rivals Meshroom’s visual quality but yields a 4.46 mm depth error, over 20$\times$ higher than Meshroom’s 0.201 mm. Conversely, on the NPHM dataset-which uses active scanning rather than photogrammetry-this systematic bias is absent, and our comparable visual quality is accurately reflected in the quantitative metrics.

Qualitative results (\autoref{fig:geom_comp}) demonstrate that our method achieves comparable visual fidelity to Meshroom and captures fine-scale geometric details, such as wrinkles and folds, that are missed or smoothed over by 2DGS and SuGaR. Additional comparison are provided in the supplemental material. 

\subsection{Ablation Studies}
\label{subsec:ablations}
\begin{figure}[t!]
    \centering
    \vspace{-5pt}
    \includegraphics[width=1\linewidth]{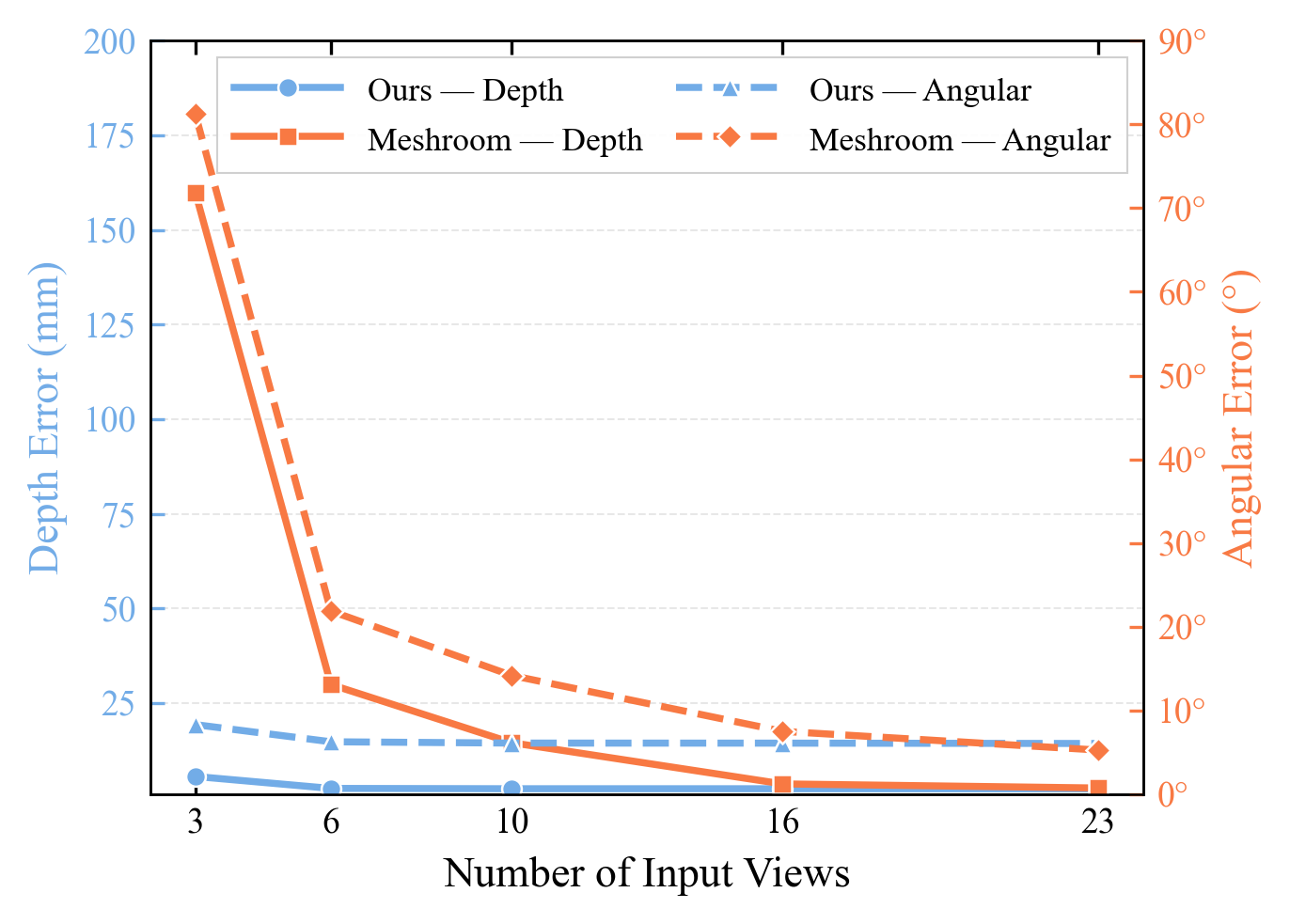}
    \vspace{-2.5em}
    \caption{Evolution of mesh reconstruction performance with the number of input camera views (NPHM dataset).}
    \label{fig:num_views}
  \vspace{-10pt}
\end{figure}

\begin{figure}
    \centering
    \includegraphics[width=1\linewidth]{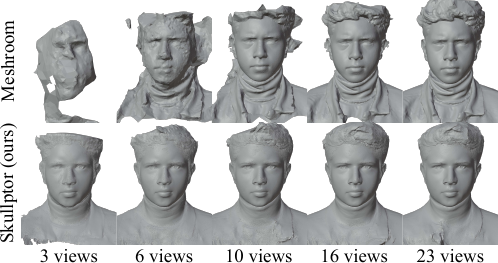}
    \caption{Visual comparison of the effect of the number of input views on reconstruction quality for Meshroom and Skullptor.}
    \label{fig:vs_meshroom}
    \vspace{-13pt}
\end{figure}

\textbf{Effect of input view count.} We evaluate our pipeline's robustness to sparse input views by comparing against Meshroom, the best performing baseline from Table~\ref{tab:geom_comp_merged}, on NPHM  using 3, 6, 10, 16, and 23 views. As shown in \autoref{fig:num_views} and \autoref{fig:vs_meshroom}, our method achieves high-fidelity reconstruction even in extremely sparse configurations of only 3 cameras, while Meshroom degrades rapidly below 16 views and fails almost completely at 3 views. This demonstrates that learned geometric priors effectively compensate for reduced view coverage, enabling practical sparse-view capture setups where traditional photogrammetry fails.

\textbf{Multi-view and cross-attention importance.}  
To isolate the architectural contributions of our model, we conducted a series of ablation experiments across four distinct configurations, summarized in Tab.~\ref{tab:combined_metrics}. We evaluate each setup on both normal estimation and the full Skullptor mesh reconstruction pipeline: (1) the original DAViD baseline; (2) the DAViD model fine-tuned on our dataset using only single-view inputs ($\textit{DAViD\_ft\_mono}$); (3) the DAViD model trained on batches of multi-view inputs but without our cross-attention mechanism ($\textit{DAViD\_ft\_multi}$); and (4) our full proposed model incorporating view-aware cross-attention with multi-view training data. 
The results of this analysis show that, first, simply fine-tuning alone on our dataset with monocular input ($\textit{DAViD\_ft\_mono}$) does not improve performance, and can even degrade it. This confirms that the improvements observed in our full model are not merely a byproduct of domain adaptation. Secondly, training on multi-view inputs of the same subjects helps the model converge to more accurate results, demonstrating that consistent predictions are fundamental to resolving geometric ambiguities. Finally, our full model with cross-attention achieves the highest performance across all benchmarks, highlighting the structural enhancements of Skullptor's normal prediction model.

\begin{table}[t]
  \centering
  \footnotesize
  \setlength{\tabcolsep}{2pt} 
  \resizebox{\columnwidth}{!}{
  \begin{tabular}{ll cc | ccc}
    & & \multicolumn{2}{c}{Normal Metrics} & \multicolumn{3}{c}{Mesh Metrics} \\
    \cmidrule(lr){2-7} 
    & Method & Ang. $\downarrow$ & Grad. $\downarrow$ & Depth $\downarrow$ & Ang. $\downarrow$ & Grad. $\downarrow$ \\
    \midrule
    \multirow{4}{*}{\rotatebox[origin=c]{90}{\scriptsize NPHM}} 
    & DAViD & 7.80 & 0.188 & \underline{2.68} & \textbf{5.68} & 0.115 \\
    & DAViD\_ft\_mono & 8.43 & \underline{0.174} & 3.53 & 6.04 & 0.116 \\
    & DAViD\_ft\_multi & \underline{7.13} & 0.176 & 2.87 & 5.98 & \underline{0.113} \\
    & Skullptor & \textbf{7.05} & \textbf{0.167} & \textbf{2.44} & \underline{5.89} & \textbf{0.110} \\
    \midrule
    \multirow{4}{*}{\rotatebox[origin=c]{90}{\scriptsize Multiface}} 
    & DAViD & 8.61 & 0.238 & 3.95 & 6.83 & 0.163 \\
    & DAViD\_ft\_mono & 8.88 & 0.228 & 4.32 & 7.06 & 0.163 \\
    & DAViD\_ft\_multi & \underline{8.59} & \underline{0.227} & \underline{3.70} & \underline{6.74} & \underline{0.161} \\
    & Skullptor & \textbf{8.54} & \textbf{0.226} & \textbf{3.29} & \textbf{6.17} & \textbf{0.156} \\
  \end{tabular}
  }
  \vspace{-0.5em}
  \caption{Comparison of normal (left) and mesh (right) error.}
  \vspace{-13pt}
  \label{tab:combined_metrics}
\end{table}

\section{Limitations and Future Work}
\label{sec:limitations}
While our method achieves high-fidelity production-quality 3D reconstructions from sparse input views, it is designed for capture setups with controlled lighting and synchronized cameras. Strong view-dependent reflections, noisy images, and facial props can cause inaccurate normal predictions that propagate to the final geometry. 
Future work includes extending our framework to jointly predict normals and albedo for complete appearance capture and incorporating material and lighting estimation to enable relighting.
%

\section{Conclusion}
\label{sec:conclusion}
We presented Skullptor, a method for high-fidelity production-quality 3D head reconstruction from sparse multi-view images that combines the complementary strengths of learning-based and optimization-based approaches. By introducing a multi-view normal prediction model that enforces geometric consistency across viewpoints and integrating it with a custom inverse rendering optimization, our approach achieves photogrammetry-level quality from fewer than 10 camera views while being an order of magnitude faster. Our results demonstrate that learned geometric priors can effectively compensate for sparse view coverage, enabling practical high-fidelity capture setups. We believe this work represents an important step toward making professional-quality 3D facial capture more accessible in production pipelines.
\newpage

{
    \bibliographystyle{ieeenat_fullname}
    \bibliography{main}

\begin{thebibliography}{69}
\providecommand{\natexlab}[1]{#1}
\providecommand{\url}[1]{\texttt{#1}}
\expandafter\ifx\csname urlstyle\endcsname\relax
  \providecommand{\doi}[1]{doi: #1}\else
  \providecommand{\doi}{doi: \begingroup \urlstyle{rm}\Url}\fi

\bibitem[tri()]{triplegangers}
Triplegangers.
\newblock \url{https://triplegangers.com/}.
\newblock Accessed: Nov. 12, 2025.

\bibitem[Alliez et~al.(2008)]{alliez2008recent}
Pierre Alliez et~al.
\newblock Recent advances in remeshing of surfaces.
\newblock \emph{Shape Anal. Struct.}, pages 53--82, 2008.

\bibitem[Arun et~al.(1987)]{arun1987least}
K~Somani Arun et~al.
\newblock Least-squares fitting of two {3-D} point sets.
\newblock \emph{IEEE Trans. Pattern Anal. Mach. Intell.}, 1987.

\bibitem[Blanz and Vetter(1999)]{blanz1999morphable}
Volker Blanz and Thomas Vetter.
\newblock A morphable model for the synthesis of {3D} faces.
\newblock In \emph{Proc. SIGGRAPH}, 1999.

\bibitem[Botsch and Kobbelt(2004)]{10.1145/1057432.1057457}
Mario Botsch and Leif Kobbelt.
\newblock A remeshing approach to multiresolution modeling.
\newblock In \emph{Proc. SGP}, 2004.

\bibitem[Botsch et~al.(2010)]{botsch2010polygon}
Mario Botsch et~al.
\newblock \emph{Polygon mesh processing}.
\newblock CRC press, 2010.

\bibitem[Brument et~al.(2024)]{Brument24}
Baptiste Brument et~al.
\newblock {RNb-NeuS}: Reflectance and {Normal-based} {Multi-View} {3D} reconstruction.
\newblock In \emph{Proc. CVPR}, 2024.

\bibitem[Bruneau et~al.(2025)]{Bruneau25}
Robin Bruneau et~al.
\newblock {Multi-view} surface reconstruction using normal and reflectance cues.
\newblock \emph{arXiv:2506.04115}, 2025.

\bibitem[Chai et~al.(2023)]{chai2023hiface}
Zenghao Chai et~al.
\newblock {HiFace}: High-fidelity {3D} face reconstruction by learning static and dynamic details.
\newblock In \emph{Proc. ICCV}, 2023.

\bibitem[Cheng et~al.(2020)]{cheng2020faster}
Shiyang Cheng et~al.
\newblock Faster, better and more detailed: {3D} face reconstruction with graph convolutional networks.
\newblock In \emph{Proc. ACCV}, 2020.

\bibitem[Dan{\v{e}}{\v{c}}ek et~al.(2022)]{danvevcek2022emoca}
Radek Dan{\v{e}}{\v{c}}ek et~al.
\newblock {EMOCA}: Emotion driven monocular face capture and animation.
\newblock In \emph{Proc. CVPR}, 2022.

\bibitem[Dib et~al.(2021{\natexlab{a}})]{dib2021practical}
Abdallah Dib et~al.
\newblock Practical face reconstruction via differentiable ray tracing.
\newblock \emph{Comput. Graph. Forum}, 2021{\natexlab{a}}.

\bibitem[Dib et~al.(2021{\natexlab{b}})]{dib2021towards}
Abdallah Dib et~al.
\newblock Towards high fidelity monocular face reconstruction with rich reflectance using self-supervised learning and ray tracing.
\newblock In \emph{Proc. ICCV}, 2021{\natexlab{b}}.

\bibitem[Dib et~al.(2022)]{dib2022s2f2}
Abdallah Dib et~al.
\newblock {S2F2}: Self-supervised high fidelity face reconstruction from monocular image.
\newblock \emph{arXiv:2203.07732}, 2022.

\bibitem[Dib et~al.(2024)]{Dib_2024_CVPR}
Abdallah Dib et~al.
\newblock {MoSAR}: Monocular semi-supervised model for avatar reconstruction using differentiable shading.
\newblock In \emph{Proc. CVPR}, 2024.

\bibitem[Dosovitskiy et~al.(2021)]{dosovitskiy2021imageworth16x16words}
Alexey Dosovitskiy et~al.
\newblock An image is worth 16x16 words: {Transformers} for image recognition at scale.
\newblock In \emph{Proc. ICLR}, 2021.

\bibitem[Egger et~al.(2020)]{Egger20Years}
Bernhard Egger et~al.
\newblock {3D} morphable face models-past, present, and future.
\newblock \emph{ACM Trans. Graph.}, 39\penalty0 (5):\penalty0 1--38, 2020.

\bibitem[Feng et~al.(2021)]{feng2021learning}
Yao Feng et~al.
\newblock Learning an animatable detailed {3D} face model from in-the-wild images.
\newblock \emph{ACM Trans. Graph.}, 40\penalty0 (4):\penalty0 1--13, 2021.

\bibitem[Furukawa et~al.(2015)Furukawa, Hern{\'a}ndez, et~al.]{furukawa2015multi}
Yasutaka Furukawa, Carlos Hern{\'a}ndez, et~al.
\newblock Multi-view stereo: A tutorial.
\newblock \emph{Found. Trends Comput. Graph. Vis.}, 9\penalty0 (1-2):\penalty0 1--148, 2015.

\bibitem[Gao et~al.(2020)]{gao2020semi}
Zhongpai Gao et~al.
\newblock Semi-supervised {3D} face representation learning from unconstrained photo collections.
\newblock In \emph{Proc. CVPRW}, 2020.

\bibitem[Giebenhain et~al.(2023)]{giebenhain2023nphm}
Simon Giebenhain et~al.
\newblock Learning neural parametric head models.
\newblock In \emph{Proc. CVPR}, 2023.

\bibitem[Giebenhain et~al.(2025)]{giebenhain2025pixel3dmm}
Simon Giebenhain et~al.
\newblock {Pixel3DMM}: Versatile screen-space priors for single-image {3D} face reconstruction.
\newblock arXiv:2505.00615, 2025.

\bibitem[Griwodz et~al.(2021)]{Meshroom}
Carsten Griwodz et~al.
\newblock {AliceVision} {Meshroom}: An open-source {3D} reconstruction pipeline.
\newblock In \emph{Proc. MMSys}, 2021.

\bibitem[Gu{\'e}don and Lepetit(2024)]{guedon2024sugar}
Antoine Gu{\'e}don and Vincent Lepetit.
\newblock {SuGaR}: Surface-aligned {Gaussian} splatting for efficient {3D} mesh reconstruction and high-quality mesh rendering.
\newblock In \emph{Proc. CVPR}, 2024.

\bibitem[Guo et~al.(2023)]{guo2023rafare}
Longwei Guo et~al.
\newblock {RaFaRe}: Learning robust and accurate non-parametric {3D} face reconstruction.
\newblock In \emph{Proc. AAAI}, 2023.

\bibitem[Hartley and Zisserman(2003)]{hartley2003multiple}
Richard Hartley and Andrew Zisserman.
\newblock \emph{Multiple view geometry in computer vision}.
\newblock Cambridge university press, 2003.

\bibitem[Hoppe et~al.(1993)]{10.1145/166117.166119}
Hugues Hoppe et~al.
\newblock Mesh optimization.
\newblock In \emph{Proc. SIGGRAPH}, 1993.

\bibitem[hsin Wuu et~al.(2023)]{wuu2023multifacedatasetneuralface}
Cheng hsin Wuu et~al.
\newblock {Multiface}: A dataset for neural face rendering.
\newblock arXiv:2207.11243, 2023.

\bibitem[Huang et~al.(2024)Huang, Yu, Chen, Geiger, and Gao]{Huang2DGS2024}
Binbin Huang, Zehao Yu, Anpei Chen, Andreas Geiger, and Shenghua Gao.
\newblock {2D} {Gaussian} {Splatting} for geometrically accurate radiance fields.
\newblock In \emph{Proc. SIGGRAPH}, 2024.

\bibitem[Jackson et~al.(2017)]{jackson2017large}
Aaron~S Jackson et~al.
\newblock Large pose {3D} face reconstruction from a single image via direct volumetric {CNN} regression.
\newblock In \emph{Proc. ICCV}, 2017.

\bibitem[Jakob et~al.(2015)Jakob, Tarini, Panozzo, and Sorkine-Hornung]{10.1145/2816795.2818078}
Wenzel Jakob, Marco Tarini, Daniele Panozzo, and Olga Sorkine-Hornung.
\newblock Instant field-aligned meshes.
\newblock \emph{ACM Trans. Graph.}, 34\penalty0 (6):\penalty0 189:1--189:15, 2015.

\bibitem[Kanazawa et~al.(2018)]{kanazawa2018learning}
Angjoo Kanazawa et~al.
\newblock Learning category-specific mesh reconstruction from image collections.
\newblock In \emph{Proc. ECCV}, 2018.

\bibitem[Ke et~al.(2025)]{ke2025marigold}
Bingxin Ke et~al.
\newblock {Marigold}: Affordable adaptation of diffusion-based image generators.
\newblock arXiv:2505.09358, 2025.

\bibitem[Kerbl et~al.(2023)Kerbl, Kopanas, Leimk{\"u}hler, and Drettakis]{kerbl3Dgaussians}
Bernhard Kerbl, Georgios Kopanas, Thomas Leimk{\"u}hler, and George Drettakis.
\newblock {3D} {Gaussian} {Splatting} for real-time radiance field rendering.
\newblock \emph{ACM Trans. Graph.}, 42\penalty0 (4), 2023.

\bibitem[Khirodkar et~al.(2024)]{khirodkar2024sapiens}
Rawal Khirodkar et~al.
\newblock {Sapiens}: Foundation for human vision models.
\newblock In \emph{Proc. ECCV}, 2024.

\bibitem[Kingma and Ba(2015)]{kingma2017adammethodstochasticoptimization}
Diederik~P. Kingma and Jimmy Ba.
\newblock {Adam}: A method for stochastic optimization.
\newblock In \emph{Proc. ICLR}, 2015.

\bibitem[Laine et~al.(2020)]{Laine2020diffrast}
Samuli Laine et~al.
\newblock Modular primitives for high-performance differentiable rendering.
\newblock \emph{ACM Trans. Graph.}, 39\penalty0 (6), 2020.

\bibitem[Lee and Lee(2020)]{lee2020uncertainty}
Gun-Hee Lee and Seong-Whan Lee.
\newblock Uncertainty-aware mesh decoder for high fidelity {3D} face reconstruction.
\newblock In \emph{Proc. CVPR}, 2020.

\bibitem[Lei et~al.(2023)]{lei2023hierarchical}
Biwen Lei et~al.
\newblock A hierarchical representation network for accurate and detailed face reconstruction from {In-The-Wild} images.
\newblock In \emph{Proc. CVPR}, 2023.

\bibitem[Li et~al.(2025)]{li2025pshumanphotorealisticsingleimage3d}
Peng Li et~al.
\newblock {PSHuman}: Photorealistic single-image {3D} human reconstruction.
\newblock arXiv:2409.10141, 2025.

\bibitem[Li et~al.(2017)]{li2017learning}
Tianye Li et~al.
\newblock Learning a model of facial shape and expression from {4D} scans.
\newblock \emph{ACM Trans. Graph.}, 36\penalty0 (6):\penalty0 194, 2017.

\bibitem[Li et~al.(2024)Li, Liu, Yan, Chen, Liang, Chen, Tan, and Long]{li2024craftsman}
Weiyu Li, Jiarui Liu, Hongyu Yan, Rui Chen, Yixun Liang, Xuelin Chen, Ping Tan, and Xiaoxiao Long.
\newblock {CraftsMan3D}: High-fidelity mesh generation with {3D} native generation and interactive geometry refiner.
\newblock \emph{arXiv:2405.14979}, 2024.

\bibitem[Lin et~al.(2020)]{lin2020towards}
Jiangke Lin et~al.
\newblock Towards high-fidelity {3D} face reconstruction from in-the-wild images using graph convolutional networks.
\newblock In \emph{Proc. CVPR}, 2020.

\bibitem[Liu et~al.(2023)]{liu2023zero1to3}
Ruoshi Liu et~al.
\newblock {Zero-1-to-3}: Zero-shot one image to {3D} object.
\newblock arXiv:2303.11328, 2023.

\bibitem[Nealen et~al.(2006)]{nealen2006laplacian}
Andrew Nealen et~al.
\newblock Laplacian mesh optimization.
\newblock In \emph{Proc. GRAPHITE}, 2006.

\bibitem[Niemeyer et~al.(2020)]{niemeyer2020differentiable}
Michael Niemeyer et~al.
\newblock Differentiable volumetric rendering: Learning implicit {3D} representations without {3D} supervision.
\newblock In \emph{Proc. CVPR}, 2020.

\bibitem[Oquab et~al.(2024)]{oquab2023dinov2}
Maxime Oquab et~al.
\newblock {DINOv2}: Learning robust visual features without supervision.
\newblock \emph{Trans. Mach. Learn. Res.}, 2024.

\bibitem[Palfinger(2022)]{palfinger2022continuous}
Werner Palfinger.
\newblock Continuous remeshing for inverse rendering.
\newblock \emph{Comput. Animat. Virtual Worlds}, 33\penalty0 (5):\penalty0 e2101, 2022.

\bibitem[Ramon et~al.(2021)]{ramon2021h3d}
Eduard Ramon et~al.
\newblock {H3D-Net}: Few-shot high-fidelity {3D} head reconstruction.
\newblock In \emph{Proc. ICCV}, 2021.

\bibitem[Ranftl et~al.(2021)Ranftl, Bochkovskiy, and Koltun]{ranftl2021visiontransformersdenseprediction}
René Ranftl, Alexey Bochkovskiy, and Vladlen Koltun.
\newblock {Vision Transformers} for {Dense Prediction}.
\newblock In \emph{Proc. ICCV}, 2021.

\bibitem[Ravi et~al.(2020)]{ravi2020accelerating3ddeeplearning}
Nikhila Ravi et~al.
\newblock Accelerating {3D} deep learning with {PyTorch3D}.
\newblock \emph{arXiv:2007.08501}, 2020.

\bibitem[Ren et~al.(2023)]{ren2023facial}
Xingyu Ren et~al.
\newblock Facial geometric detail recovery via implicit representation.
\newblock In \emph{Proc. FG}, 2023.

\bibitem[Rosu and Behnke(2023)]{rosu2023permutosdf}
Radu~Alexandru Rosu and Sven Behnke.
\newblock {PermutoSDF}: Fast multi-view reconstruction with implicit surfaces using permutohedral lattices.
\newblock In \emph{Proc. CVPR}, 2023.

\bibitem[Saleh et~al.(2025)]{saleh2025david}
Fatemeh Saleh et~al.
\newblock {DAViD}: Data-efficient and accurate vision models from synthetic data.
\newblock arXiv:2507.15365, 2025.

\bibitem[Sch\"{o}nberger and Frahm(2016)]{schoenberger2016sfm}
Johannes~Lutz Sch\"{o}nberger and Jan-Michael Frahm.
\newblock {Structure-from-Motion} revisited.
\newblock In \emph{Proc. CVPR}, 2016.

\bibitem[Sch\"{o}nberger et~al.(2016)]{schoenberger2016mvs}
Johannes~Lutz Sch\"{o}nberger et~al.
\newblock Pixelwise view selection for unstructured {Multi-View} stereo.
\newblock In \emph{Proc. ECCV}, 2016.

\bibitem[Shi et~al.(2023)]{shi2023zero123plus}
Ruoxi Shi et~al.
\newblock {Zero123++}: a single image to consistent multi-view diffusion base model.
\newblock arXiv:2310.15110, 2023.

\bibitem[Sivanandan and Liscio(2017)]{EvaScanner}
Janujah Sivanandan and Eugene Liscio.
\newblock Assessing structured light {3D} scanning using {Artec} {Eva} for injury documentation during autopsy.
\newblock 2017.

\bibitem[Surazhsky and Gotsman(2003)]{:10.2312/SGP/SGP03/020-030}
Vitaly Surazhsky and Craig Gotsman.
\newblock Explicit surface remeshing.
\newblock In \emph{Proc. SGP}, 2003.

\bibitem[Szeliski(2022)]{szeliski2022computer}
Richard Szeliski.
\newblock \emph{Computer vision: algorithms and applications}.
\newblock Springer Nature, 2022.

\bibitem[Taubner et~al.(2024)]{taubner2024cap4dcreatinganimatable4d}
Felix Taubner et~al.
\newblock {CAP4D}: Creating animatable {4D} portrait avatars.
\newblock arXiv:2412.12093, 2024.

\bibitem[Wang et~al.(2018)]{wang2018pixel2mesh}
Nanyang Wang et~al.
\newblock {Pixel2Mesh}: Generating {3D} mesh models from single {RGB} images.
\newblock In \emph{Proc. ECCV}, 2018.

\bibitem[Wang et~al.(2024)]{wang20243d}
Zidu Wang et~al.
\newblock {3D} face reconstruction with the geometric guidance of facial part segmentation.
\newblock In \emph{Proc. CVPR}, 2024.

\bibitem[Wei et~al.(2024)]{wei2024meshlrm}
Xinyue Wei et~al.
\newblock {MeshLRM}: Large reconstruction model for high-quality meshes.
\newblock \emph{arXiv:2404.12385}, 2024.

\bibitem[Wu et~al.(2024)]{wu2024unique3dhighqualityefficient3d}
Kailu Wu et~al.
\newblock {Unique3D}: High-quality and efficient {3D} mesh generation.
\newblock arXiv:2405.20343, 2024.

\bibitem[Yang et~al.(2024{\natexlab{a}})]{depth_anything_v1}
Lihe Yang et~al.
\newblock Depth anything: Unleashing the power of large-scale unlabeled data.
\newblock In \emph{Proc. CVPR}, 2024{\natexlab{a}}.

\bibitem[Yang et~al.(2024{\natexlab{b}})]{depth_anything_v2}
Lihe Yang et~al.
\newblock Depth anything {V2}.
\newblock \emph{arXiv:2406.09414}, 2024{\natexlab{b}}.

\bibitem[Ye et~al.(2024)]{ye2024stablenormal}
Chongjie Ye et~al.
\newblock {StableNormal}: Reducing diffusion variance for stable and sharp normal.
\newblock \emph{ACM Trans. Graph.}, 2024.

\bibitem[Zheng et~al.(2022)]{zheng2022farl}
Yinglin Zheng et~al.
\newblock General facial representation learning in a visual-linguistic manner.
\newblock In \emph{Proc. CVPR}, 2022.

\end{thebibliography}
}
\newpage



%
%
%

\setcounter{section}{0}
\setcounter{figure}{0}
\setcounter{table}{0}
\renewcommand{\thesection}{S\arabic{section}}
\renewcommand{\thefigure}{S\arabic{figure}}
\renewcommand{\thetable}{S\arabic{table}}

\twocolumn[
  \begin{center}
    {\Large \bf Skullptor: High Fidelity 3D Head Reconstruction in Seconds\\ 
    with Multi-View Normal Prediction \par}
    \vspace{10pt}
    {\Large Supplementary Material \par}
    \vspace{20pt}
  \end{center}
]

\section{Additional Results}
\label{sec:results}
Additional qualitative results for our method, Skullptor, including 4D sequences and extended comparisons with state-of-the-art methods for normal estimation and mesh reconstruction, are available in the \href{https://ubisoft-laforge.github.io/character/skullptor/}{supplemental webpage}.
 
\subsection{Impact of the Normal Estimator}
To validate the importance of our multi-view normal estimator (Sec.~\ref{method_normal}), we replace it with several monocular baselines (Sapiens 0.3B/2B, DAViD) within our full pipeline. For these monocular methods, we independently predict normals for each of the 10 input views and then run the same inverse rendering optimization to recover the geometry. Since monocular predictors process each view independently without enforcing multi-view consistency, they produce normals that can be geometrically inconsistent across viewpoints, leading to degraded reconstruction quality. 
As shown in \autoref{tab:ablation_normal_estimator}, our multi-view estimator consistently produces superior geometry. Notably, while Sapiens 2B performed well on the normal comparison evaluation (\autoref{tab:normal_comp_merged}), its final geometry is poor (6.76 mm error on NPHM). We attribute this to its tendency to produce overly smooth normals, an observation supported by the high average normal gradient error it achieved in \autoref{tab:normal_comp_merged}, which leads to a loss of fine surface details during optimization. This discrepancy demonstrates that multi-view processing during the normal prediction stage translates to more accurate and coherent final geometry.

\begin{table}[!t]
\centering
\resizebox{\columnwidth}{!}{
\begin{tabular}{@{}llccc@{}}
\toprule
Dataset & Normal Estimator & Depth (mm) $\downarrow$ & Angular Err. $\downarrow$ & Avg. Normal Grad. Err. $\downarrow$ \\
\midrule
\multirow{4}{*}{\rotatebox[origin=c]{90}{Multiface}} 
& Sapiens 0.3B & 5.28 & 8.75 & 0.177 \\
& Sapiens 2B & 3.79 & 6.76 & 0.167 \\
& DAViD & \underline{3.74} & \underline{6.73} & \underline{0.163} \\
\cmidrule{2-5}
& Skullptor (Ours) & \textbf{3.20} & \textbf{6.45} & \textbf{0.157} \\
\midrule\midrule
\multirow{4}{*}{\rotatebox[origin=c]{90}{NPHM}} 
& Sapiens 0.3B & 6.38 & 8.23 & 0.124 \\
& Sapiens 2B & 6.76 & 7.91 & 0.121 \\
& DAViD & \underline{2.65} & \textbf{5.91} & \underline{0.116} \\
\cmidrule{2-5}
& Skullptor (Ours) & \textbf{2.39} & \underline{6.07} & \textbf{0.112} \\
\bottomrule
\end{tabular}
}
\caption{Ablation study on normal estimator impact on final mesh quality (10 views).}
\label{tab:ablation_normal_estimator}
\end{table}

\begin{figure}[t]
    \includegraphics[width=\columnwidth]{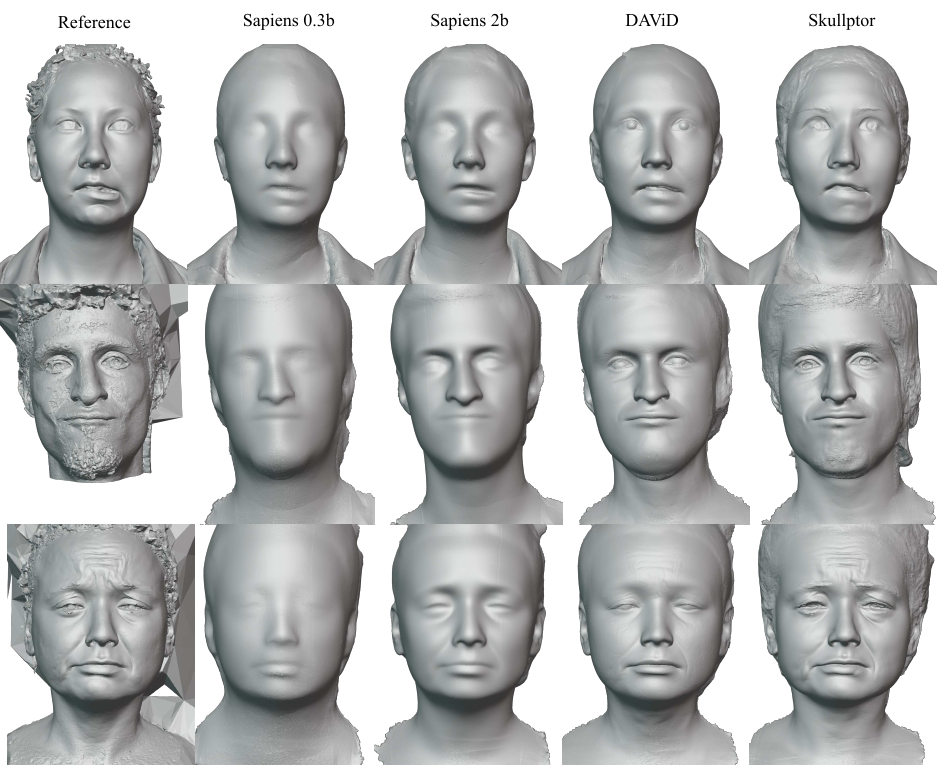}
    \caption{Qualitative impact of normal estimation quality on mesh reconstruction.}
    \label{fig:mesh_predictor_comp_compressed}
\end{figure}

To complement these quantitative findings, we provide visual comparisons in \autoref{fig:mesh_predictor_comp_compressed} illustrating how different normal estimators affect the remeshing stage. Monocular predictors exhibit noticeable identity drift and loss of structural fidelity in the reconstructed mesh due to the lack of shared geometric information during view processing. In contrast, our multi-view approach maintains high-frequency details and identity consistency by leveraging cross-view features during the initial estimation phase.

\subsection{Generalization to In-the-Wild Capture}  
\begin{figure}[t]
  \centering
  \vspace{-10pt}
  \begin{minipage}{\linewidth}
    \centering
    \includegraphics[width=\linewidth]{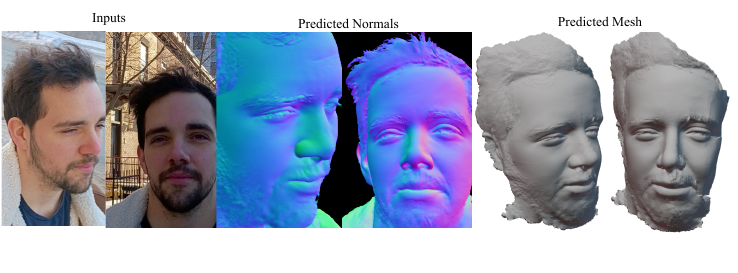}
    \vspace{-26pt} 
    \caption{Skullptor results on 12 phone-captured images.}
    \label{fig:in_the_wild}
  \end{minipage}
  \vspace{-10pt}
\end{figure}

Although our primary evaluation focuses on lightstage capture setups, DAViD's backbone provides robust priors from 300K diverse images that our fine-tuning specializes rather than overwrites. As a result, Skullptor generalizes over four distinct capture setups (TripleGangers, Multiface, NPHM, and ours) with varying camera configurations, lighting, and subjects. Additionally, we include an in-the-wild result in Fig.~\ref{fig:in_the_wild}, suggesting potential for generalizing beyond controlled environments.

\section{Implementation Details For Multi-View Normal Prediction}
In this section, we provide implementation details for the normal prediction (Sec.\ \ref{method_normal} in the main paper).


\subsection{Training Dataset}
Our training dataset is curated from high-quality Triplegangers \cite{triplegangers} facial scan assets to ensure diversity across ethnicity, age, and gender. Each scan in the original dataset was captured using a professional light stage setup comprising 55 synchronized high-end cameras, followed by a photogrammetry reconstruction pipeline with manual cleaning to ensure geometrical accuracy. Each subject is captured while performing 20 static facial expressions, resulting in 1,000 unique expression instances across the dataset. The distribution of subjects in our training dataset is given in Table~\ref{tab:triplegangers_demographics}.
\begin{table}[h!]
    \centering
    \caption{Subject distribution in the Triplegangers training dataset.}
    \begin{tabular}{@{}lcc@{}}
        \toprule
        & \textbf{Male} & \textbf{Female} \\
        \midrule
        Number of subjects & 18 & 32 \\
        Age range & 18--64 & 18--75 \\
        Mean age & 37.2 & 42.8 \\
        \midrule
        Age $<$ 25 & 9 (50\%) & 14 (44\%) \\
        Age 25--65 & 8 (44\%) & 12 (38\%) \\
        Age $>$ 65 & 1 (6\%) & 6 (19\%) \\        
        \bottomrule
    \end{tabular}
     \label{tab:triplegangers_demographics}
\end{table}

\noindent \textbf{Training-validation split.} We reserve 5 subjects (100 expressions) for validation and use the remaining 45 subjects (900 expressions) for training. This subject-level split ensures that the model's generalization is evaluated on entirely unseen identities.

\noindent \textbf{Mesh normalization.} To standardize the input geometry across subjects and expressions, we apply Procrustes analysis using 3D facial landmarks to align each scan to a normalized template facial mesh. Specifically, we center each mesh at the origin and scale it to fit within a unit sphere. This normalization ensures consistent scale and orientation across all training samples.
Our foreground segmentation allows us to get the full upper part and neck, not only the face. 

\noindent \textbf{Rendering setup.} For every subject's expression, we render both RGB images and rasterized surface normal maps from 48 virtual cameras at $512 \times 512$ resolution using nvdiffrast \cite{Laine2020diffrast}. The rasterized normal maps are transformed into each camera’s local coordinates by multiplying them with the rotation of its model-view matrix. Camera viewpoints are randomly sampled for every expression-subject pair to maximize viewpoint diversity during training. In total, the training set consists of $43 200$ images.

\subsection{Camera View Generation}
As mentioned above, we generate 48 diverse camera views per expression for every subject. The camera viewpoints are selected by randomly sampling from a parameterized camera distribution that ensures comprehensive coverage of the facial region. For the 48 cameras associated with an expression, we construct a camera transformation matrix using the following random sampling of parameters as listed in \autoref{tab:camera_projection_params}. The camera transformation matrix is constructed by combining rotations around the X-axis (pitch) and Y-axis (yaw), followed by translation and scaling operations. Random sampling allows the model to learn from a wide range of viewing angles and distances.

\begin{table}[h]
\centering
\caption{Camera Sampling and Projection Parameters}
\begin{tabular}{ll}
\toprule
\multicolumn{2}{c}{\textbf{Camera Sampling}} \\
\midrule
Pitch angle $\theta_{\text{pitch}}$ & $\mathcal{U}(-35^\circ, 35^\circ)$ \\
Yaw angle $\theta_{\text{yaw}}$     & $\mathcal{U}(-90^\circ, 90^\circ)$ \\
Initial radius $\mathbf{r}$ & 2.0 \\
Extra 3D translation $\mathbf{t}$          & $\mathcal{U}(-0.2, 0.2)^3$ \\
Scale factor $z$                      & $\mathcal{U}(1.0, 1.8)$ \\
\midrule
\multicolumn{2}{c}{\textbf{Projection Matrix}} \\
\midrule
Image resolution $w = h$             & 512 pixels \\
Focal lengths $f_x = f_y$            & 512 pixels \\
Principal point $c_x = c_y$          & 256 pixels \\
Depth range (near $n$, far $f$)     & $0.001$ -- $1000$ \\
\bottomrule
\end{tabular}
\label{tab:camera_projection_params}
\end{table}
\subsection{Training Procedure}
Each epoch iterates over all training expressions, selecting six random viewpoints from the 48 pre-rendered cameras for each expression. This sampling strategy encourages the model to integrate information across diverse view combinations, reducing the risk of overfitting to particular camera configurations.


We use the AdamW optimizer with differential learning rates: $1 \times 10^{-4}$ for the newly added multi-view attention components and $1 \times 10^{-5}$ for the DAViD backbone \cite{saleh2025david}. 
A weight decay of $1 \times 10^{-4}$ is used while training the model for 100 epochs with a batch size of 6 views per sample. The model is trained on an NVIDIA Quadro RTX 8000 GPU using PyTorch, with a total training time of 82 hours on 900 training expressions (45 subjects) and 100 validation expressions (5 subjects).
\subsection{Inference}
Our architecture extends the monocular foundation model \cite{saleh2025david} with a multi-view cross-attention mechanism to aggregate information across different input viewpoints in each iteration. Although the model is trained exclusively with 6-view batches, it generalizes seamlessly to arbitrary numbers of views during inference.
This generalization capability stems from the inherent properties of cross-attention. Each target view attends to all available context views through learned query-key-value projections. The key and value matrices encode information from each context view into a fixed embedding dimension, which the attention mechanism aggregates regardless of the number of input views. During training with randomly sampled 6-view subsets, the model learns to flexibly combine information from diverse viewpoint configurations, making it naturally robust to different numbers of views at test time.

At inference, the model accepts any number of input views (from 3 to 26 in our experiments) without architectural modifications or retraining. The cross-attention operation dynamically adjusts to the available views, computing attention weights that reflect the geometric consistency of each viewpoint for reconstructing the target normal map. 

\section{Implementation Details For Mesh Optimization}
In this section, we provide implementation details for the Mesh optimization (Sec.~\ref{ssec:mesh_optim}).

For optimization, we first pre-compute the per-pixel weight matrix from the normal maps (obtained from our normal prediction model) using Eq.\ 9 of the main paper. The target normal maps are then masked using the foreground segmentation model from~\cite{saleh2025david} to retain the complete head region, including the face, neck, and upper torso.

We use the optimization framework from Continuous Remeshing~\cite{palfinger2022continuous} to refine the mesh geometry. The optimization proceeds for 300 steps with a learning rate of 0.3. At each iteration, we rasterize the normal maps from the current mesh state using nvdiffrast (Eq. 6), compute the normal loss (Eq. 8) along with the Laplacian regularization term (Eq. 7) with $\lambda_{\mathrm{lap}} = 0.1$), and update the mesh vertices. At the end of each optimization step, we apply adaptive remeshing operations from~\cite{palfinger2022continuous}. These operations consist of edge splits, collapses, and flips that dynamically adjust mesh resolution to local geometric complexity while preventing degeneracies such as collapsed faces and self-intersections.

\subsection{Initialization}
For the optimization to converge successfully, the initial sphere must be positioned close to the target mesh location in 3D space. To estimate this location, we leverage the known camera extrinsics. We cast rays from each camera center through its optical axis and compute the point in 3D space that minimizes the sum of distances to all rays. This point serves as the center of the initial sphere. While this approach is most accurate when all cameras are focused on the same region, it remains robust even when camera viewing directions vary, provided the cameras roughly face a common area of interest.

\begin{figure*}[t]
    \centering
    \includegraphics[width=1\linewidth]{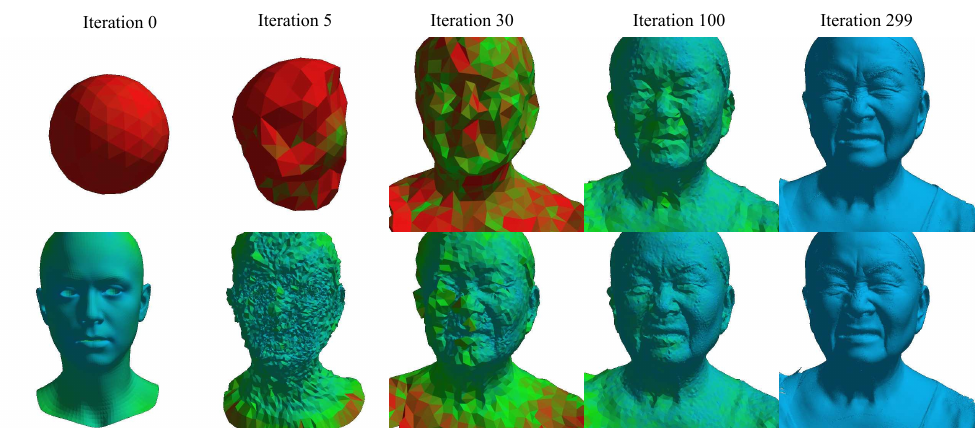}
    \caption{Effect of initialization on mesh optimization. Vertex color indicates local edge length (Red short large edges, blue for small edges). \textbf{Top:} Sphere initialization. \textbf{Bottom:} Template mesh initialization. Facial characteristics from the template mesh are discarded, and both initializations converge to comparable results in 300 iterations.}
    \label{fig:sphere_vs_template}
\end{figure*}

During optimization, we observed that initializing from a neutral facial template rather than a sphere does not improve convergence or final quality (Figure~\ref{fig:sphere_vs_template}). Both converge to comparable results in 300 iterations.

 \subsection{Mesh Resolution} 
 The optimization controls granularity by specifying the maximum vertex count and minimum edge length. We observed no quality gains beyond 500k vertices, suggesting resolution ceases to be a limiting factor at that scale. Skullptor's evaluation used this vertex count, which is the same order of magnitude as 2DGS/SuGaR ($\sim$450k) and Meshroom ($\sim$250k).
\section{Evaluation Protocol on NPHM and Multiface}
Because the NPHM dataset~\cite{giebenhain2023nphm} provides only raw textured meshes without corresponding camera images, we generate a synthetic multi-view dataset by rendering each mesh from 23 virtual cameras. The cameras are uniformly sampled on a sphere around the head to provide full $360$ degrees of face coverage, with the head centered at the origin. We render $2048\times2048$ color images and corresponding camera-space normal maps using NVDiffRast, without shading (texture only). These renders form the input to all reconstruction baselines. High image resolution is necessary for fair comparison, as Meshroom~\cite{Meshroom}, 2DGS~\cite{Huang2DGS2024}, and SuGaR~\cite{guedon2024sugar} exhibit degraded performance at lower input resolutions.

As Multiface~\cite{wuu2023multifacedatasetneuralface} provides original lightstage-captured images at $1334\times2048$ resolution, we directly use the real multi-view RGB frames as input during evaluation. The dataset contains synchronized multi-view video sequences of subjects performing a variety of expressions. From the camera array, we select 26 viewpoints that provide full facial coverage and feed their captured images to all reconstruction methods.

For our method, the input images are processed to match the $512 \times 512$ training resolution of our multi-view normal predictor. For NPHM, we downsample the $2048 \times 2048$ rendered images. For Multiface, we resize and pad the $1334 \times 2048$ captured images to obtain $512 \times 512$ inputs. All baselines operate directly on the original image resolution ($2048\times2048$ for NPHM renders and the native resolution for Multiface captures of $1334\times2048$).

To evaluate reconstruction quality, we measure performance on novel viewpoints not used during reconstruction. We render 12 additional frontal-facing cameras, explicitly excluding any back-of-head viewpoints, to test generalization to unseen poses. For each method, we render normal and depth maps from its reconstructed mesh at $512\times512$ from these novel viewpoints and compare them against ground-truth renderings from the corresponding ground-truth mesh under identical camera poses.

To ensure consistency across all methods, every reconstructed mesh is first aligned to a canonical coordinate system using Procrustes analysis (as described in Sec.\ \ref{ssec:mesh_optim}). Depth and normal comparisons are performed after this canonical alignment for all methods, which is necessary because some baselines may output meshes in arbitrary coordinate frames.

Evaluation focuses on the frontal facial region, which is the primary area of interest for head-level fidelity. We extract per-view facial masks using Facer~\cite{zheng2022farl}, removing hair, neck, and clothing. These masks are applied both to predictions and ground truth to ensure that metrics reflect facial geometry alone.
\end{document}